\pdfoutput=1

\documentclass[11pt]{article}

\usepackage{ACL2023}

\usepackage{times}
\usepackage{latexsym}

\usepackage[T1]{fontenc}

\usepackage[utf8]{inputenc}

\usepackage{microtype}

\usepackage{inconsolata}

\usepackage{graphicx}
\usepackage{tabularx}
\usepackage{amsmath,bm}
\usepackage{amssymb}
\usepackage{multirow}
\usepackage{makecell}
\usepackage[toc,page]{appendix}

\title{Learning to Reduce: Optimal Representations of Structured Data \\ in Prompting Large Language Models}

\author{Younghun Lee$^{\dagger}$, Sungchul Kim$^{\ddagger}$, Tong Yu$^{\ddagger}$, Ryan A. Rossi$^{\ddagger}$, Xiang Chen$^{\ddagger}$ \\
       $^{\dagger}$Department of Computer Science, Purdue University \\
       $^{\ddagger}$Adobe Research\\
        \texttt{younghun@purdue.edu}; \texttt{\{sukim,tyu,ryrossi,xiangche\}@adobe.com}\\}

\begin{document}
\maketitle
\begin{abstract}
Large Language Models (LLMs) have been widely used as general-purpose AI agents showing comparable performance on many downstream tasks. However, existing work shows that it is challenging for LLMs to integrate structured data (e.g. KG, tables, DBs) into their prompts; LLMs need to either understand long text data or select the most relevant evidence prior to inference, and both approaches are not trivial.
\\
In this paper, we propose a framework, Learning to Reduce, that fine-tunes a language model to generate a reduced version of an input context, given a task description and context input. The model learns to reduce the input context using On-Policy Reinforcement Learning and aims to improve the reasoning performance of a fixed LLM. Experimental results illustrate that our model not only achieves comparable accuracies in selecting the relevant evidence from an input context, but also shows generalizability on different datasets. We further show that our model helps improve the LLM’s performance on downstream tasks especially when the context is long.
\end{abstract}

\section{Introduction}

State-of-the-art Large Language Models (LLMs) such as GPT-4 \citep{gpt4} have shown the ability to understand language on many downstream tasks, and they exhibit satisfactory performance compared to models that are fine-tuned on the specific tasks \citep{wei2022chain, kojima2022large, wang2022self, yu2022generate}. 

Despite their ability, LLMs find it challenging to integrate structured data into their prompts. Structured data such as knowledge graphs, tables, and databases, have structural dependencies among entities and instances which is distinct from natural language input. 
To help LLMs better understand such structure, existing research shows the effectiveness of simple linearization methods that convert knowledge graph triplets and table cells into natural language \citep{yasunaga-etal-2021-qa, tan2023make, hegselmann2023tabllm}.

Another aspect that makes it difficult to prompt LLMs with structured data is the lengthy context; knowledge graphs and tables often contain more than a few hundred entities and relations. The maximum input sequence length of the recent LLMs keeps increasing, yet one cannot put faith in the performance of LLMs when the context is long. \citet{liu2023lost} shows that the performance of the ChatGPT \citep{chatgpt} on a multi-document QA task drops more than $20\%$ in the following circumstances: when the context is lengthy, and when the key information is located in the middle of the prompt. This implies that LLMs are likely to perform worse on structured data because most structured data tend to be lengthy as well.

To avoid such issues, \citet{jiang2023structgpt} prompt the ChatGPT to retrieve the most relevant evidence from the structured data in order to better perform on QA tasks. Unfortunately, this is not a trivial process; experimental results show that around $70\%$ and $30\%$ of the errors come from incorrect data selection in KGQA tasks and table QA tasks, respectively.

In this paper, we explore a method to improve LLMs' reasoning ability on structured data by Learning to Reduce; the model efficiently reduces the input context data by identifying the relevant evidence to the downstream tasks. We hypothesize that prompting LLMs with long context data requires careful data selection, yet there is a limited number of studies on this approach. We specifically focus on structured data QA tasks, mainly on tables, as they not only bring a long context problem, but also introduce more challenges regarding the structure. We train a model to generate a reduced context using On-Policy Learning given the question and the original context.

We empirically show that our model shows comparable performance in reducing the input context on a widely used table QA dataset, WikiTableQuestions \citep{pasupat-liang-2015-compositional}. Additionally, our model outperforms other baseline models including the most recent GPT-4 \cite{gpt4} on the robustness test on unseen data distribution. Lastly, we argue that the output of our model helps LLMs perform more accurately on downstream QA tasks, especially when the context is lengthy.

\begin{figure*}[t!]
    \centering
    \includegraphics[width=0.98\textwidth]{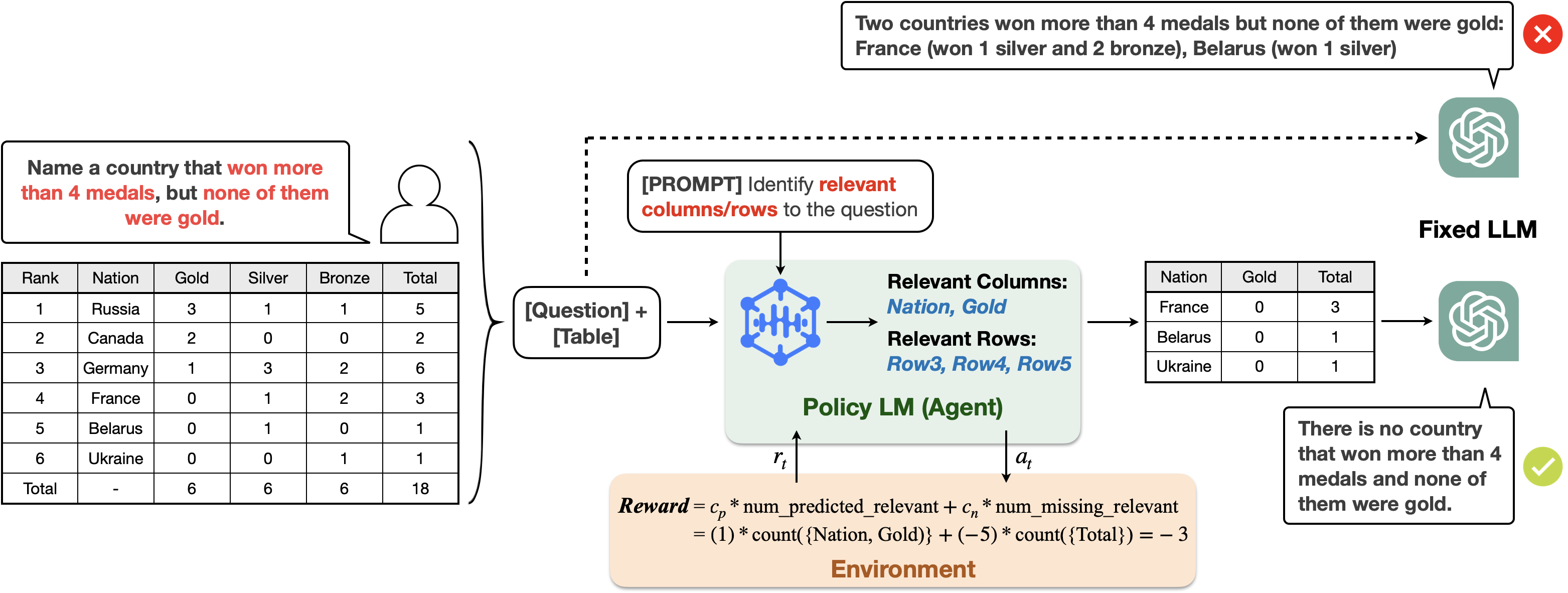}
    \caption{Comparison between original prompting (dotted arrow path) and Learning to Reduce (solid arrow path). Given a pair of input question and context, a language model (blue hexagon) learns a policy to generate the relevant rows and columns by getting rewards. The model gets a high negative reward $c_n$ for the number of necessary items that are missing from the prediction. By updating the policy network parameters with rewards, our model generates the correct reduced context which leads the fixed LLM model (e.g. GPT-4) to perform more accurately on downstream tasks.}
    \label{fig:model}
\end{figure*}

\textbf{Key Contributions}: To the best of our knowledge, this is the first attempt to train a language model to reduce the input context of structured data QA tasks. With better model training, we expect our model could be used as a pre-prompting tool for any structured data QA tasks. This would ultimately maximize the reasoning ability of LLMs as well as the cost efficiency of using them.

\section{Approach}
In this paper, we mainly focus on better utilizing the most recent LLM, GPT-4\footnote{We used the version gpt-4-0613}, in solving QA tasks on tabular data.

\subsection{Reducing Input Context}
Using LLMs to solve QA tasks on tabular data often encounters a maximum input sequence length issue; it is very likely to reach the maximum sequence length in prompting LLMs when there are a few hundred rows in the table. 

One of the dominant approaches used in previous work on table QA is text-to-SQL; models are trained to generate a correct SQL query to answer the given question. The major drawback of this approach is that it inherently lacks reasoning abilities and fails to answer questions that require additional context information \citep{chen-etal-2020-hybridqa}. Alternatively, researchers tried selecting the most relevant rows and columns before performing question answering. The results from StructGPT \citep{jiang2023structgpt} illustrate that this method introduces selection errors as much as QA reasoning errors. 



\subsection{Training Pipeline}

The aforementioned observation motivates the necessity of a robust and accurate way to select the most relevant evidence prior to prompting LLMs. We design a training pipeline for a language model that learns to generate the relevant evidence for a given QA instance. 

Formally, we define an input space that consists of an input context $\bm{c}$, a task description $\bm{x}$, and an output $\bm{y}$. In the table QA task, for instance, $\bm{c}$ is an input table, $\bm{x}$ is an input question, and $\bm{y}$ is an answer to the question. We train a language model, $\theta(\bm{z|x,c})$, that learns to generate a reduced input context $\bm{z}$. As our final objective is to help LLMs better answer the questions, we prompt a LLM with reduced context to generate an output. Denoting $\psi_{LLM}$ as a fixed LLM model used for question answering, the final output $\bm{y}$ follows $\bm{y} \sim \psi_{LLM}(\cdot | \bm{x, z})$

\subsection{LM as a Policy Network}
\label{sec:language-model-as-policy-network}
We consider a language model as a policy network. Following an approach proposed by \citet{li2023guiding}, we first fine-tune the language model parameters on the dataset prior to applying policy learning objectives. 

Before fine-tuning, we need a dataset that has relevant rows and columns annotated for input questions and tables. We use a table QA dataset which has a text-to-SQL annotation to automatically identify the relevant evidence. The general idea of annotation is executing SQL queries while iteratively removing rows and columns from the input tables; if executing a SQL query can generate an answer on the table even after removing some rows and columns, then the removed items are irrelevant to the input question.

\begin{table}\footnotesize
\centering
\begin{tabularx}{0.95\columnwidth}{X|r}
\hline
\multicolumn{2}{c}{\textbf{WTQ with valid SQUALL annotation}} \\
\hline
\# of train / valid / test set & 10K / 3K / 1.5K \\
\# of unique tables & 5.4K \\
\# of questions per table & 2.73 \\
\hline
Avg / Max \# of columns & 9 / 25\\
Avg / Max \# of rows & 46 / 753\\
Avg / Max \# of cells & 230 / 3,832\\
\hline
Avg / Max \# of context tokens & 1,333 / 20,324\\
\# instances > 4,096 tokens & 248\\
\# instances > 8,192 tokens & 60\\
\hline
\end{tabularx}
\caption{\label{tab:data-statistics}
Statistics of the WTQ dataset with valid SQUALL annotation
}
\end{table}

Using the WikiTableQuestions (WTQ) dataset \citep{pasupat-liang-2015-compositional} and its corresponding text-to-SQL annotations from SQUALL \citep{shi-etal-2020-potential}, we identify relevant rows and columns by iteratively removing them one by one from the original tables until the annotated SQL queries fail to generate the answers. 
Table \ref{tab:data-statistics} describes the statistics of the data.

We use a sequence-to-sequence language model with FLAN-T5-Large \citep{chung2022scaling} checkpoint as a base model. For better optimization, we train two separate models, one for the column reduction and the other for the row reduction. The column reduction model gets an input that consists of a prompt, an input question, and a list of column headers, and it is fine-tuned to generate a sequence of relevant columns. For the row reduction model, we represent row items indicating column names and the corresponding cell values as in \verb|Row0: (Col0,Val0), (Col1,Val1)|. We represent row items only with the relevant columns, assuming that the columns are already reduced with $100\%$ accuracy. This is mainly to make the input as concise as possible.



\subsection{RL Agent Training}
We consider a fine-tuned language model as an initial policy network parameters and further optimize the model by maximizing the rewards. The policy network gets positive rewards for selecting the correct relevant items, and gets two types of negative rewards for incorrect selection. 

Having irrelevant rows and columns as relevant evidence is not preferable but the error is not critical; when the LLMs use the output of the policy network later, LLMs can still perform a QA inference with the table containing a few irrelevant rows and columns. In this case, the policy network gets a small negative reward.

Another type of negative reward is given when the policy network fails to generate the relevant rows and columns. In this case, the LLMs would have to perform inference without the necessary information. We consider this as a critical error and the policy network gets a very high negative reward. 

\subsection{Model Architecture}
\label{appendix:model}
The initial policy network parameters are pre-trained language model that is small enough to fine-tune. We use a sequence to sequence language model with FLAN-T5-Large \citep{chung2022scaling} checkpoint as a base model and fine-tune the model with the annotation of relevant rows and columns for each question and table pair.
Mathematically, we fine-tune the language models by maximizing the log-likelihood as follows:

$$\mathcal{L}_{FT} = -\mathbb{E}\text{log}\theta_{LM}(\bm{z|x,c})$$

We further tune the language model by computing the rewards from the model's predictions on relevant rows and columns. The model gets positive rewards for selecting the correct relevant items, and gets negative rewards for incorrect selection. The model is penalized more when it does not select the relevant items compared to when it selects irrelevant items. To formally define the parameter update condition, we aim to maximize the following objective:
$$\text{max}_{\theta_{LM}}\mathbb{E}_{\bm{z}\sim \theta_{LM}(\cdot|\bm{x,c})}[\mathcal{R}(\bm{x,z})]$$

To make the optimization tractable for policy network, we employ proximal policy optimization (PPO) method \citep{schulman2017proximal}. We consider a fine-tuned language model as an initial policy network (i.e., $\pi_0=\theta_{LM}$) and update the policy network $\pi$ using PPO. The policy network's generation of relevant information can be considered as a Markov Decision Process $\langle \mathcal{S}, \mathcal{A}, r, \mathcal{P} \rangle$, where $\mathcal{S}$ is a state space, $\mathcal{A}$ is an action space, $r$ is a reward function, and $\mathcal{P}$ is transition probabilities. At each time step $t$ in an episode, the model selects an action (i.e., generating tokens of relevant information) from $\mathcal{A}$, based on the state of the current time step. The state at time $t$ is defined with the input and the policy network's previous generations, that is, $\bm{z_{t}} = \pi(\cdot|\bm{x, z_{<t-1}})$. The episode ends when the policy network generates an end-of-sequence token.

Following the existing RL approaches on NLP applications \citep{ziegler2019fine}, we employ KL divergence penalty rewards that dynamically adapt the coefficient $\beta$ in different time steps to minimize excessive parameter updates from the initial policy. The action space for token generation often stretches to the size of vocabulary and it makes optimization costly. To minimize such issues, \citep{ramamurthy2022reinforcement} proposed NLPO, an approach that masks out the least probable tokens using top-$p$ sampling. We set $p$ as 0.9 in the experiments. The policy network's reward function $r\bm{(x, z)}$ is defined as:

$$r(\bm{x,z}) = \mathcal{R}(\bm{x,z}) - \beta\log{\frac{\pi(\bm{z|x,c})}{\theta(\bm{z|x,c})}}$$
$$\bm{e}_t = \text{CLIP}\left(\frac{\text{KL}(\pi_t, \theta)-\text{KL}_{\text{target}}}{\text{KL}_{\text{target}}} , -0.2, 0.2 \right)$$
$$\beta_{t+1} = \beta_t(1+K_{\beta}\bm{e}_t)$$


\begin{table}[t]\footnotesize
\centering
\begin{tabularx}{0.95\columnwidth}{X|r|r}
\hline
\multicolumn{3}{c}{\textbf{Accuracy on Cotext Reduction}}\\
\hline
\hline
\textbf{Column Reduction} & \textbf{WTQ} & \textbf{HybQA} \\
\hline
RoBERTa Sequence Clf & 92.87 & 70.19\\
Zero-shot GPT-4 & 74.03 & 71.48\\
Fine-tuned FLAN-T5 & 90.19 & 82.72\\
$\pmb{\star}$ Learning to Reduce (\textit{ours}) & \textit{91.82} & \textbf{\textit{87.22}}\\
\hline
\hline
\textbf{Row Reduction} & \textbf{WTQ} & \textbf{HybQA} \\
\hline
RoBERTa Sequence Clf & 96.46 & 93.56\\
Zero-shot GPT-4 & 86.13 & 92.12\\
Fine-tuned FLAN-T5 & 95.21 & 90.22\\
$\pmb{\star}$ Learning to Reduce (\textit{ours}) & \textit{96.17} & \textbf{\textit{93.78}}\\
\hline
\end{tabularx}
\caption{\label{tab:context-reduce-experiments}
Input context reduction tested on an in-distribution test set (WTQ) and on an unseen test set (HybQA).
}
\end{table}

\section{Models and Experiments}
This section describes the baseline models we implement as well as the performance of these models compared to our proposed approach.

\subsection{Baseline Models}
First, we fine-tune a RoBERTa sequence classifier \citep{liu2019roberta} that learns whether the input question and each column/row item are aligned. For instance, the input to the classifier is \verb|<col1>[SEP]<Question>| and the output is either 0 or 1 based on whether the \verb|col1| is relevant to the \verb|Question|. 

Another baseline is a zero-shot GPT-4 model. In this setting, we provide an input question and a table, then ask the GPT-4 model to select relevant rows and columns to answer the question.

Lastly, we use a fine-tuned sequence-to-sequence language model as a baseline. This is to measure the effectiveness of the On-Policy Learning component in context reduction.

\subsection{Performance and Generalizability}
We use recall values as a metric for context reduction and measure how many relevant rows and columns are selected from the models. Table \ref{tab:context-reduce-experiments} shows the experimental results. 

The models are tested on the WTQ test set as well as on Hybrid QA \citep{chen-etal-2020-hybridqa}. Hybrid QA is a table QA task that not only requires reasoning on a given table but also needs world knowledge. We test the models on Hybrid QA to evaluate their generalizability. The model needs to be robust to different data distribution because fine-tuning is not always possible. Thus we argue that the model's performance on Hybrid QA should be considered more significantly in comparing different models. 

RoBERTa baseline performs well on context reduction on the WTQ test set. However, the model's performance significantly drops when it is tested on Hybrid QA; the column reduction model scores $93\%$ recall on WTQ while $70\%$ on Hybrid QA. This implies that this baseline model needs to be fine-tuned again for different datasets, which is not ideal.


GPT-4 is a very large and highly contextual model, yet it has a hard time understanding the relatedness between a table and a question. This result aligns with the selection error statistics in the existing research \citep{jiang2023structgpt}. As the model is not fine-tuned on a specific dataset, its performance on WTQ and Hybrid QA are similar as expected. A slight drop from $74\%$ to $71.5\%$ implies that Hybrid QA is a little more challenging dataset than WTQ.

Regarding the language model approaches, On-Policy Learning helps improve the performance of the language model. Not only does the Learning to Reduce show comparable performance on the WTQ test set, but it exhibits remarkable results on Hybrid QA. Our proposed model achieves $87\%$ recall on the Hybrid QA dataset and outperforms other non-LM baselines by at least $22\%$. 
This result implies that policy networks are able to capture more general knowledge about context reduction. Although it does not beat the baseline tested on WTQ, we argue that our model outperforms other baselines because of its generalizability.





\begin{figure}[t!]
    \centering
    \includegraphics[width=0.98\columnwidth]{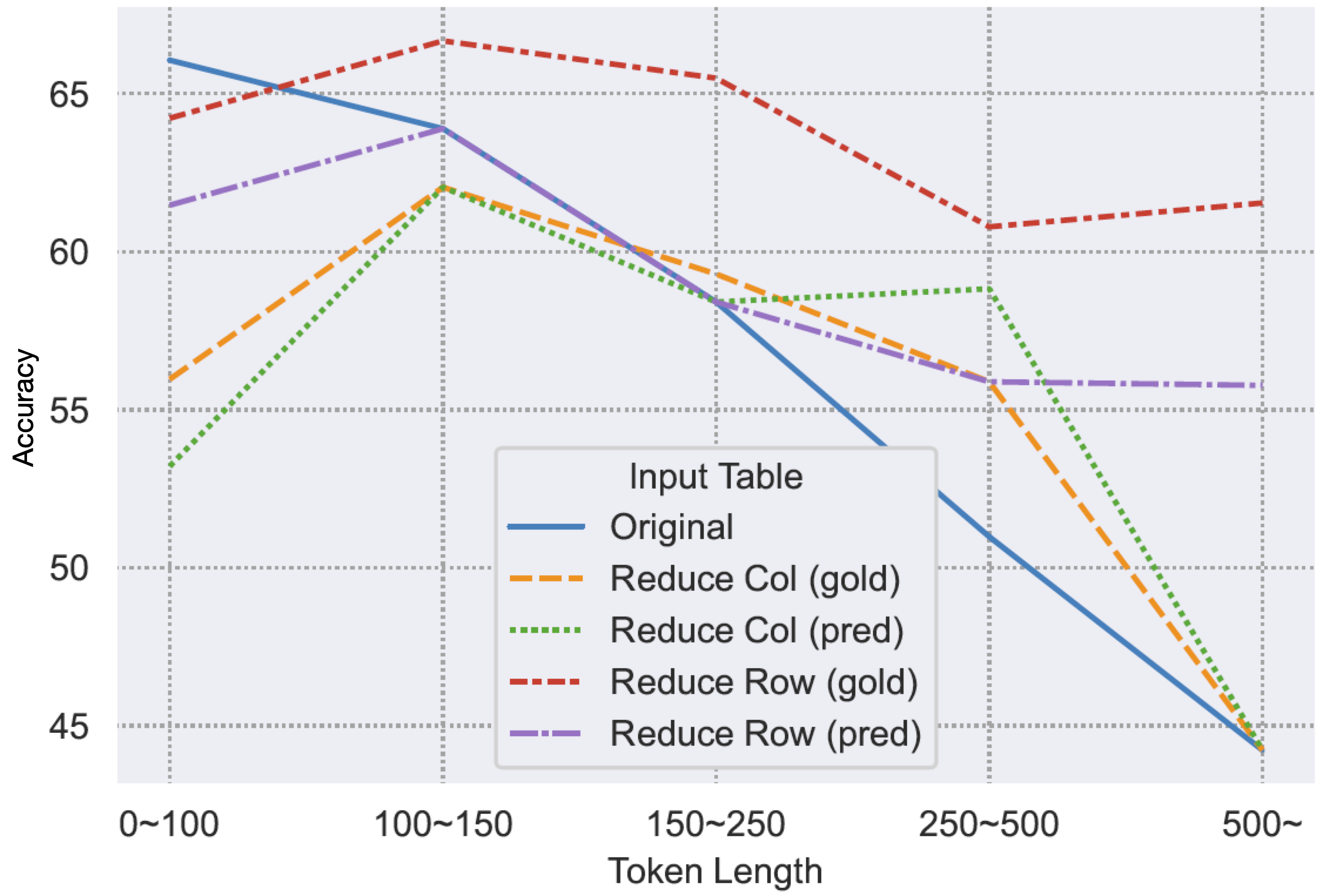}
    \caption{Accuracy of GPT-4 model on WTQ test set with different input context tables. Reducing both rows and columns (red, purple) is more powerful when the context is longer.
    }
    \label{fig:downstream-task}
\end{figure}

\subsection{Downstream Task Performance}
The ultimate goal of the model is to help LLMs better perform on downstream tasks. We use a GPT-4 model as a fixed QA reasoner while changing the input. 
Figure \ref{fig:downstream-task} illustrates the accuracy of the GPT-4 model in answering the WTQ questions when prompted with different context inputs. 

The accuracy of the original input consistently drops as the table gets longer (blue line in Figure \ref{fig:downstream-task}), while the performance of the reduced input context is significantly more stable, especially when it is longer (red and purple line). 
The purple line shows that our proposed method provides better QA results compared to pure LLM-based approaches (blue), even though the policy network's recall on context reduction is not 100\%. This result supports our hypothesis that a lengthy input context creates difficulty for LLMs to perform inference. Our proposed method proves to be more impactful for maximizing LLMs' ability in understanding longer structured data.

\section{Related Work}
\textbf{Structured Data LLM}: Recently, various approaches have been proposed in integrating structured data into LLM prompts. \citet{ye2023large} proposed a framework DATER which uses LLMs to decompose a pair of input questions and a table into a sub-table and a set of sub-questions, then executes queries on the sub-questions to improve the reasoning ability of the LLM. \citet{hegselmann2023tabllm} suggested TabLLM which first serializes feature names and values from the tables into natural language and then fine-tunes LLMs on downstream tasks. StructGPT \citep{jiang2023structgpt} iteratively runs a reading-then-reasoning approach on structured data QA tasks and gets the most relevant evidence from knowledge graphs, tables, and databases prior to prompting LLMs. 

\textbf{LLM Prompt Engineering}: Another pathway to improve the downstream task performance of a fixed LLM is to tune the prompts. One of the approaches is to optimize the instruction text in the prompt. GrIPS \citep{prasad-etal-2023-grips} uses a gradient-free, edit-based instruction search method. 
AutoPrompt \citep{shin-etal-2020-autoprompt} and RLPrompt \citep{deng-etal-2022-rlprompt} train the model to generate secret recipe tokens that help LLMs better perform on downstream tasks. Directional Stimulus Prompting \citep{li2023guiding} works on a similar approach but they train the model to generate keywords or hints that can directly help LLMs to accomplish the downstream tasks.

\section{Discussion and Conclusion}
More context reduction experiments on different structured data QA datasets will be conducted in future work. Also, the LLM's inference results on downstream tasks can be integrated as additional reward signals to better lead the model to generate reduced context.

In this paper, we propose Learning to Reduce, a language model learns to generate a reduced context through On-Policy Learning. Trained on a table QA dataset, WikiTableQuestions, our model takes a question and a table as input, and outputs a set of relevant rows and columns to answer the question. Our model not only shows comparable performance in correctly reducing the input context, but shows generalizability on an unseen dataset. We further show that the output of our model improves the LLM's performance on downstream tasks especially when the context is lengthy. 


\section*{Ethics Statement}
To the best of our knowledge, this work has not violated any code of ethics.


\bibliography{custom}
\bibliographystyle{acl_natbib}

\appendix

\section{Experiment Details}
\label{appendix:experiments}
All models are trained and tested on 8-core NVIDIA Tesla A10 GPU with 24GB RAM. Policy networks are trained over 10 iterations and evaluated for every 3 iterations. The total amount of policy network training took 18 hours. 

Language models are prompted as \textit{``Select relevant columns from a table to answer a question. Output `@' if done generating. Question: \{Table QA question\}, List of column headers: \{Column headers\}''}. When there are hundreds of rows in the table and representing them exceeds the maximum token limit of the language model, we truncate the table and prompt the model. Thus in some cases, the language model can generate only the end-of-sequence token when the given sub-table does not contain the relevant evidence.

For the generalizability test, we manually annotate 200 instances of the Hybrid QA test set in order to identify the relevant rows and columns for a given question and table pair.

\end{document}